\documentclass[runningheads]{llncs}

\usepackage[T1]{fontenc}
\usepackage{amsmath}

\usepackage{graphicx}
\usepackage{multirow}
\usepackage[table,xcdraw]{xcolor}

\usepackage{lscape}
\usepackage{lineno}
\usepackage[utf8]{inputenc} 
\usepackage[T1]{fontenc}    
\usepackage{hyperref}       
\usepackage{url}            
\usepackage{booktabs}       
\usepackage{amsfonts}       
\usepackage{nicefrac}       
\usepackage{microtype}      
\usepackage[utf8]{inputenc}
\usepackage{amsmath}
\usepackage{amssymb}

\usepackage{stfloats}

\usepackage{bm}
\usepackage[algoruled, longend, boxed]{algorithm2e}
\usepackage{mathrsfs}
\usepackage{float}
\usepackage{subfigure}
\usepackage{siunitx}
\usepackage{xcolor}
\DeclareUnicodeCharacter{00A0}{~}
\nolinenumbers

\SetKwInput{KwInput}{Input}                
\SetKwInput{KwOutput}{Output}              
\SetKwInput{KWInitialization}{Initialization}                
\SetKwInput{KWReturn}{Return}                

\begin{document}
\title{Multi-modal learning for predicting the genotype of glioma\thanks{Under review}}
\titlerunning{Multi-modal learning for predicting the genotype of glioma}
\author{Yiran Wei\inst{1} \and Xi Chen\inst{2} \and Lei Zhu\inst{3} \and Lipei Zhang\inst{4} \and   Carola-Bibiane Schönlieb\inst{4} \and Stephen J. Price\inst{1}   \and  Chao Li\thanks{Corresponding author}\inst{1,4}} 
%
\authorrunning{Y. Wei et al.}
%
\institute{Department of Clinical Neurosciences, University of Cambridge  
\and Department of Computer Science, University of Bath
\and Department of Electronic and Computer Engineering, The Hong Kong University of Science and Technology
\and Department of Applied Mathematics and Theoretical Physics, University of Cambridge }
\maketitle              
\begin{abstract}
The isocitrate dehydrogenase (IDH) gene mutation is an essential biomarker for the diagnosis and prognosis of glioma. It is promising to better predict glioma genotype by integrating focal tumor image and geometric features with brain network features derived from MRI. Convolutions neural networks show reasonable performance in predicting IDH mutation, which, however, cannot learn from non-Euclidean data, e.g., geometric and network data. In this study, we propose a multi-modal learning framework using three separate encoders to extract features of focal tumor image, tumor geometrics and global brain networks. To mitigate the limited availability of diffusion MRI, we develop a self-supervised approach to generate brain networks from anatomical multi-sequence MRI. Moreover, to extract tumor-related features from the brain network, we design a hierarchical attention module for the brain network encoder. Further, we design a bi-level multi-modal contrastive loss to align the multi-modal features and tackle the domain gap at the focal tumor and global brain. Finally, we propose a weighted population graph to integrate the multi-modal features for genotype prediction. Experimental results on the testing set show that the proposed model outperforms the baseline deep learning models. The ablation experiments validate the performance of different components of the framework. The visualized interpretation corresponds to clinical knowledge with further validation. In conclusion, the proposed learning framework provides a novel approach for predicting the genotype of glioma.

\end{abstract}
\section{INTRODUCTION}

Glioma is the most common malignant brain tumor in adults with remarkable heterogeneity and diverse survival outcomes~\cite{li2019intratumoral,li2019decoding,li2019low}. The mutation of the isocitrate dehydrogenase (IDH) gene is one of the most significant molecular markers for the diagnosis and prognosis of glioma~\cite{louis20162016}. 
The current gold standard of IDH mutation detection, i.e., immunohistochemistry and targeted gene sequencing, is invasive and time-consuming, hindering timely clinical decision making~\cite{louis20162016}. 

An increasing number of studies have shown that MRI can predict the IDH mutation. Compared to radiomics approaches, deep learning has achieved better performance~\cite{liang2018multimodal}. However, most deep learning models are based on convolutional neural networks (CNN), which cannot leverage the information offered by other non-Euclidean data modalities. Recent studies show that the geometric data describing tumor shape provide robust tumor phenotyping across multiple tissue histology and imaging modalities. In addition, glioma tends to invade the whole brain beyond the focal tumor. Characterizing the global brain  using the network approach has shown significance in predicting survival and cognitive decline in brain tumor patients~\cite{liu2020altered,wei2021structural}. Hence, integrating multi-modal data, including tumor image, tumor geometrics, and global brain network, could enhance glioma genotype prediction. 


Multi-modal learning shows excellent performance in integrating multi-modal data and minimizing the domain gap between modalities. For example, cross-modal attention is shown able to align fine-grained features between different modalities~\cite{lee2018stacked}. Additionally, cross-modal contrastive loss shows promising performance in extracting global representations from image and corresponding texts~\cite{zhang2020contrastive}. 
Nonetheless, existing approaches are designed for data modalities with one-to-one correspondence, which may not suit the data modalities with inclusion relation, e.g., tumor images with localized information of focal tumor and brain networks containing the information from both focal tumor and global brain.


This study develops a learning framework that generates multi-modal data, extracts, and integrates multi-modal features for boosting glioma genotype prediction. 
Specifically, apart from the image and geometric data produced from the tumor segmentation masks, we design a self-supervised approach to generate brain networks from anatomical MRIs. 
Then, we design three separate encoders for multi-modal feature extraction to characterize glioma from different aspects. At the same time, a hierarchical attention module is specially designed for the brain network encoder to assist the feature extraction. 
After that, a bi-level multi-modal contrastive loss is designed to tackle the inclusion-relation domain gap between the focal tumor and the global brain. Finally, we construct a weighted population graph approach that models the patient cohort as a large graph based on multi-modal features.A GNN is trained to classify nodes on the population graph to predict IDH mutation of patients.
Our contributions include:
\begin{itemize}
    \item Structural brain networks are conventionally constructed from diffusion MRI. To mitigate the limited availability of diffusion MRI, we propose a self-supervised approach to reconstruct the edge attributes of the brain network from anatomical MRI through contrastive representative learning, which could help transfer the knowledge of diffusion MRI to anatomical MRI. 

    \item We design a hierarchical attention module that sequentially attends to the edges and nodes of the brain network for identifying the brain network features associated with the focal tumor. This approach could allow the brain network encoder to extract the most relevant brain network features and reduce the confounding effect from  concomitant pathology.
    
    \item We present a bi-level contrastive loss for multi-modal data, which aligns the tumor-level features from the focal tumor image and geometric points cloud and then aligns the tumor-level features with the brain-level network features. This approach could reflect the gradient tumor infiltration and tackle the domain gap across the focal tumor and the global brain. 
    
   \item We construct a population graph for modelling the patient cohort with multi-modal data. The weighted nodes represent the multi-modal features of individual patients, while the weighted edges represent the continuous similarity between patients. This approach could help better integrate multi-modal features and characterise the patient cohort.

\end{itemize}

\section{Related work}

\subsection{Genotype prediction}

The studies of predicting glioma genotypes consist of radiomics-based machine learning methods and deep learning methods. The radiomics-based machine learning approaches first extract hand-crafted features from the tumor core. Feature selection is performed before training models for predicting the genotype~\cite{bhandari2021noninvasive}. For example, Gihr~\textit{et al.} successfully used intensity-based radiomics features to predict IDH mutation with reasonable accuracy~\cite{gihr2020histogram}. However, the reproducibility and generalizability of radiomics are often limited by the non-standard feature engineering and selection procedure. 

The end-to-end deep learning models, i.e., ResNet, DenseNet, provide a more robust prediction for tumor genotype over radiomics approaches~\cite{ahmad2019predictive,liang2018multimodal}. Liang~\textit{et al.} used a 3D-DensNet to predict the IDH mutation, establishing the feasibility of CNN predicting glioma genotype~\cite{liang2018multimodal}. Other deep learning models incorporate radiomics features into the model. Choi~\textit{et al.} integrated radiomics features into the later layers of CNN to enhance prediction~\cite{choi2021fully}, which outperforms the conventional ResNet. 
Despite achieving reasonable performance, the CNN-based models may not learn the information encompassed in the non-Euclidean data, e.g., geometric points cloud and brain networks, which provide crucial tumor biology and neuroscience information. Hence, we propose specialized encoders to obtain features from multi-modal data.

\subsection{Structural brain networks in glioma}

Structural brain network is a graph representation of the complex connectivity among brain regions~\cite{bullmore2011brain}, where the nodes represent the brain regions, defined according to neuroanatomy, and the edges represent the white matter connections among the regions. 
To generate structural brain networks, most studies use the approaches based on the diffusion MRI, which promises to indicate subtle tumor invasion \cite{wei2021quantifying,wei2021structural,van2017subventricular}. However, a robust model training is significantly limited by the data availability of the diffusion MRI. Recent studies indicate that the scalar map of diffusion MRI can be successfully generated from a single anatomical T1 sequence~\cite{gu2019generating}, which suggests the high-level correlation between anatomical MRI and diffusion MRI, indicating the potential of constructing brain networks using anatomical MRI. However, a single T1 sequence is insufficient to characterize the heterogeneous structural alternation caused by glioma invasion. Therefore, we proposed reconstructing edge attributes by transferring the knowledge of diffusion MRI to multi-sequence MRI using a contrastive loss. Studies of diffusion-based brain networks generally only include edge attributes. To characterize the brain regions 
invaded by glioma, we further develop an autoencoder approach to reconstruct node attributes based on regional multi-sequence MRI.

\subsection{Multi-modal learning}

Multi-modal learning is the deep learning approach that learns from more than one data modality, e.g., images, text, points cloud. Multi-modal learning has shown promising performance in a series of learning schemes. 
Lee~\textit{et al.} proposed a stacked cross attention to discover the full latent alignments between image regions and words in a sentence. Through inferring image-text similarity, the model produced interpretable prediction results~\cite{lee2018stacked}. Zhang~\textit{et al.} employed a contrastive loss between the lung X-ray and corresponding medical reports to extract relevant representations from both images and text~\cite{zhang2020contrastive}. 
Nevertheless, existing methods are not designed for data modalities with inclusion relation, e.g., focal tumor and global brain. Therefore, we propose a bi-level contrastive loss to align the features from the focal tumor and global brain levels.

\subsection{Graph neural networks}
The fast-developing graph neural network (GNN) family promises to extract features and learning from the geometric data, e.g., points cloud, which can be readily reconstructed from MRI~\cite{qi2017pointnet++}. For example, Qi~\textit{et al.} proposed hierarchically generating a graph of points cloud and recursively trained a GNN, which effectively learned local features from the geometric points cloud of the objects. 

Further, brain networks are naturally learnable by the GNN due to the graph format. Based on brain networks, GNN has shown high performance in classifying diseases. Ma~\textit{et al.} proposed a combination of recurrent neural network and GNN with an attention-guided random walk module to extract longitudinal structural graph features from the brain network for patient classification~\cite{ma2020attention}. The results showed that the attention mechanism could reveal the most critical brain regions and temporal domain during AD progression. 
Nonetheless, the attention mechanism designed for other diseases may not suit glioma due to the distinct pathophysiology. We thus develop a hierarchical attention module that could attend to the brain structure to reduce the confounding effect from concomitant pathology and capture tumor-specific features.

Finally, GNN also shows high performance in classifying the nodes in a large graph such as citation networks~\cite{cabanes2013cora}. 
The capability of GNN in handling large graphs could be transferred to patient classification tasks. Parisot \textit{et al.} proposed a population graph to model the dementia cohort by regarding imaging features of individual patients as nodes, while the clinical similarity between patients as edges~\cite{parisot2018disease}. A GNN is trained to classify patients, outperforming traditional machine learning models, e.g., random forest. This study develops a population graph to integrate the multi-modal features. Additionally, we permute the edge and node weights to select the best combination in constructing the population graph. 

\subsection{Differences from conference papers}
This study is the extension of our two previous papers in four aspects ~\cite{wei2021predicting,wei2022collaborative}. Firstly, for brain network reconstruction, we propose a contrastive learning approach to replace the original autoencoder for the brain network edge reconstruction, which additionally incorporates the knowledge from diffusion MRI. Secondly, we combine brain networks with focal tumor data (images and geometrics) to comprehensively characterize glioma. Thirdly, we design an attention module and bi-level multi-modal contrastive loss to extract the most relevant features from the multi-modal data.  Finally, we construct a population graph for feature integration and patient classification.

\section{METHODOLOGY}
\subsection{Study overview}
\begin{figure*}[t]
\centerline{\includegraphics[width=\textwidth]{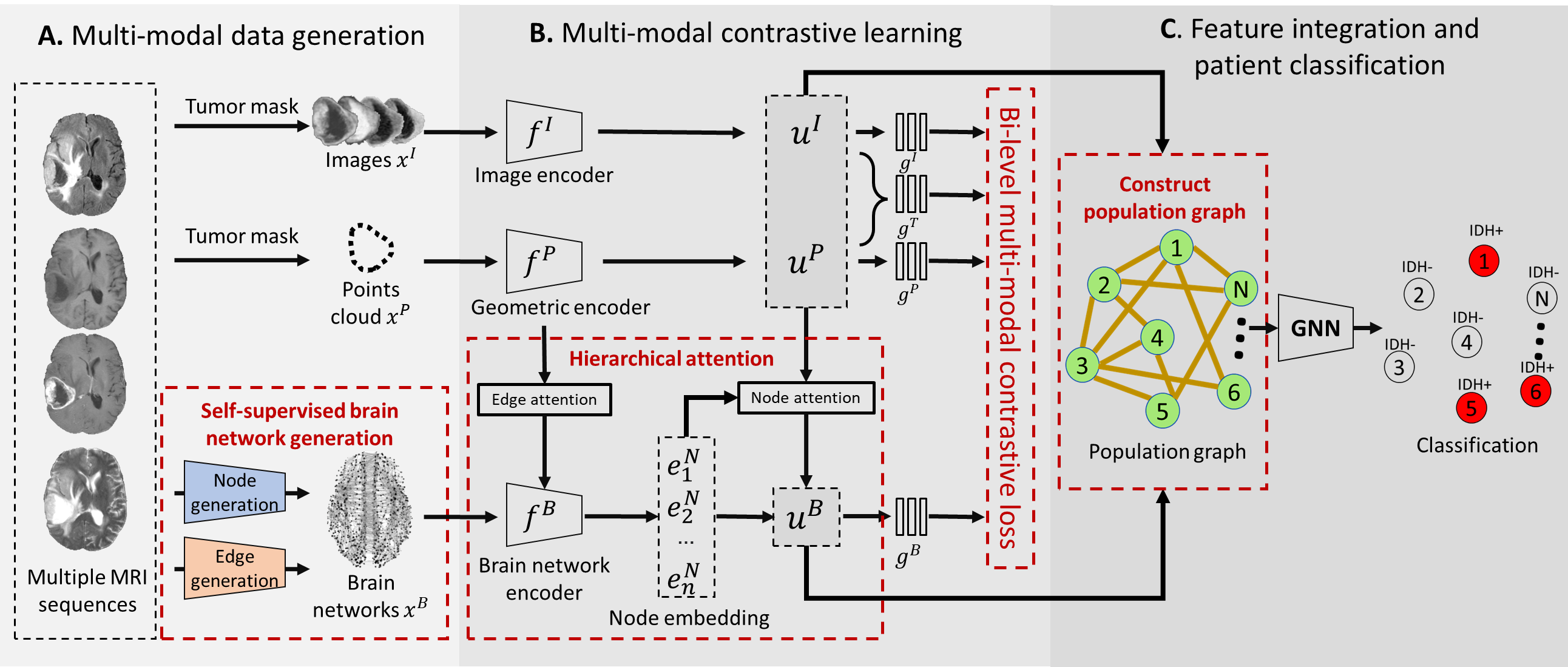}}
\caption{Study overview: A. Multi-modal data generation. Image $x^I$ and geometric $x^P$ data are generated from tumor masks, while brain network data $x^B$ are generated from the pretrained self-supervised models. B. Features of tumor image ($u^I$), tumor geometrics ($u^P$), focal tumor($\langle u^{P}, u^{I} \rangle$) and global brain network ($u^B$) are projected by respective projection heads ($g^I$, $g^P$, $g^T$ and $g^B$) for bi-level contrastive learning. A hierarchical attention module 
attend to the edges and nodes in the brain network. C. A population graph is used to integrate multi-modal features and classify patients using a GNN. }
\label{Fig:design}
\end{figure*}

As shown Fig. \ref{Fig:design}, our glioma genotype prediction network has three stages: (1) generating multi-modal data of tumor image, tumor geometrics and brain networks from the multi-sequence MRI; (2) multi-modal contrastive learning extracting features from both focal tumor image, tumor geometrics and global brain networks; (3) feature integration to construct a population graph for patient classification and genotype prediction.

\subsection{Multi-modal data generation}

\begin{figure}[t]
\centerline{\includegraphics[width=0.9\columnwidth]{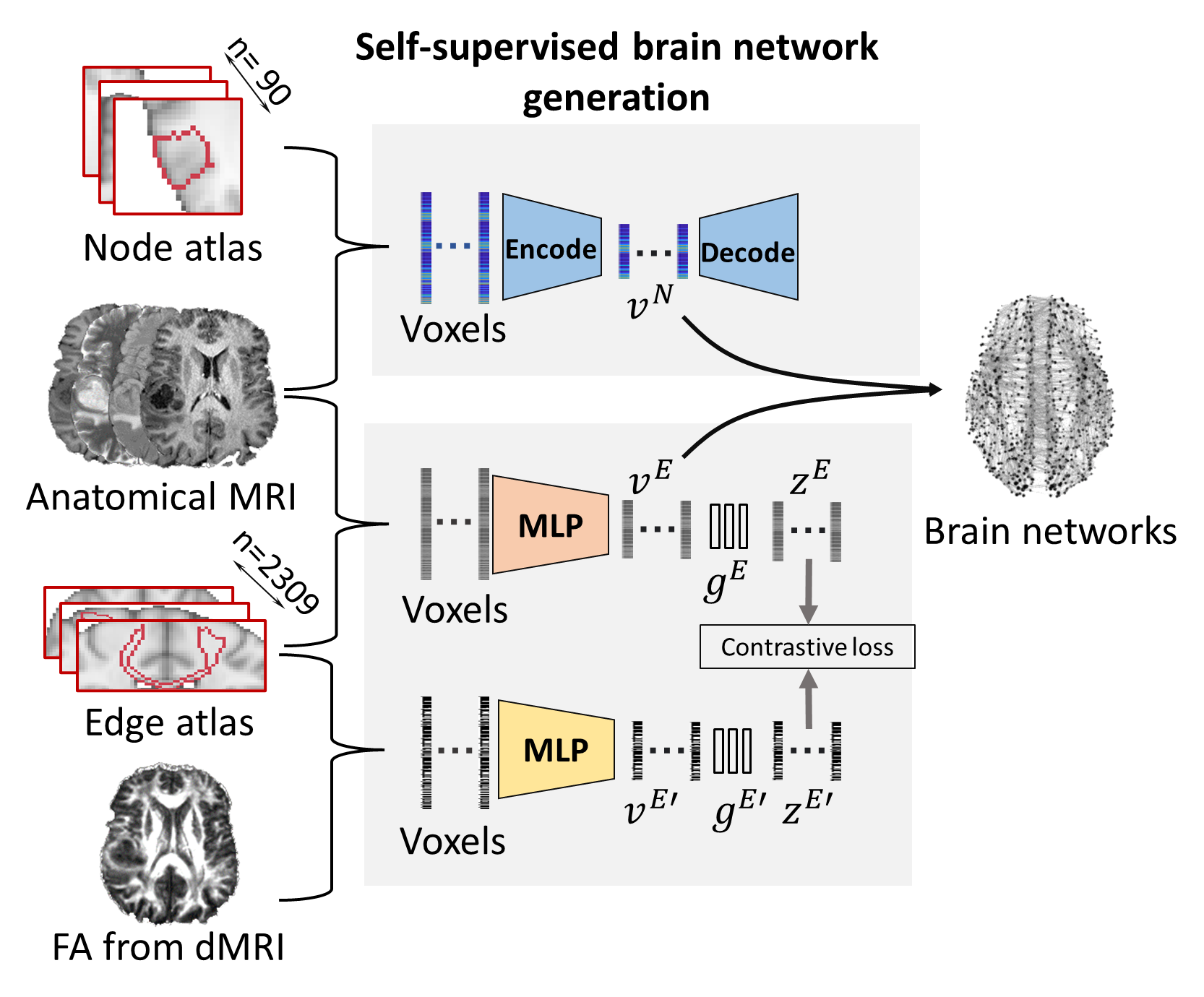}}
\caption{Brain network generation. Two self-supervised models are trained to extract node/edge attributes ($v^N$,$v^E$) from node/edge atlas bounded MRI voxels: Node attributes are extracted by the autoencoder, while edge attributes are reconstructed through contrastive learning between anatomical MRI and FA map of dMRI using projection head ($g^{E}$,$g^{E'}$), projected latent features ($z^{E}$,$z^{E'}$) and a contrastive loss.}
\label{Fig:autoencoder}
\end{figure}

Our method starts by generating three data modalities from the input multiple MRI sequences (see Fig. \ref{Fig:design}A), and the three data modalities are: (1) the image data of focal tumor (denoted as $x^{I}$) is obtained by assigning Boolean values on the tumor masks and the MRI; (2) the tumor geometric data (denoted as $x^{P}$), in the form of points cloud, is generated by sampling the surface meshes of tumor masks using a standard farthest point sampling strategy; and (3) the brain networks (denoted as $x^{B}$) is generated by two self-supervised neural networks (NNs) detailed below. 

\subsubsection{Brain networks construction via self-supervised NNs}
The brain networks, consisting of reconstructed nodes and edges, are generated based on a prior neuroanatomy atlas. As shown in the upper half of Fig~\ref{Fig:autoencoder}, voxels enclosed by the 90 cortical/ subcortical brain regions on the atlases~\cite{tzourio2002automated} are extracted and fed into an NN-based autoencoder (AE) to produce the node attributes $u^{N}$ of the brain networks. The AE consists of a NN encoder that extracts high-level representation vectors from the voxels in the brain region and a NN decoder that attempts to restore 
the voxels from the representation vectors. By adopting this self-supervised model, representations of the voxels in the brain regions could be extracted as node attributes. 

We use the probabilistic tractography atlas as the regions of interest for reconstructing edge attributes of brain networks ~\cite{wei2021quantifying}, indicating the 2,309 pathways of white matter tracts connecting the 90 brain regions. Due to the clinical significance of the fractional anisotropy (FA) map derived from the diffusion MRI in characterizing brain connectivity, we utilize the FA map to guide the edge attributes extraction of anatomical MRI. Firstly, voxels of anatomical MRI and the corresponding FA map enclosed by the tractography atlas are input into two multilayer perceptron (MLP), which respectively extract the attribute vectors $v^{E}$ and $v^{E'}$ from voxels.
Next, two projection heads $g$ and $g'$ project the attribute vectors to a common latent space, where domain alignment is performed between the latent attributes of anatomical MRI ($z^E=g(v^{E}_n)$) and FA ($z^{E'}=g'(v^{E'}_n)$) using a contrastive loss. The edge attributes extracted from the anatomical MRI contain corresponding information in the FA map. The contrastive loss of the edge $\mathcal{L}_{edge}$ is defined as:
\begin{equation}\label{equ:brainnetedge}
       \mathcal{L}_{edge}= \frac{1}{M} \sum_{i = 1}^{M} (-\log \frac{\exp(S(z^E_{n},z^{E'}_{n})/\tau)}{\sum_{m\neq n}^{M} \exp(S(z^E_{n},z^{E'}_{m}))/\tau)}) \ ,
\end{equation}
where $n$ is the target tract index, while $m$ is the index of other tracts in the minibatch; $S(\cdot)$ is the similarity score; $\tau$ is the temperature parameter; $M$ is the size of the minibatch. 

Finally, the node and edge attributes reconstructed from the pre-trained models are reformatted into the brain network data $x^{B}=\{ v^{E}, v^{N} \}$.

\subsection{Multi-modal learning for image, geometrics and brain networks}
The proposed multi-modal learning framework extracts features from the three modalities data, i.e., focal tumor image, focal tumor geometric and global brain networks. Moreover, hierarchical attention is developed for the brain network encoder to extract tumor-related brain network features. Finally, the extracted features are projected into a shared latent space for bi-level multi-modal contrastive learning, which could minimize the domain gap from the tumor level (image and geometrics) across the global brain level (focal tumor and brain networks). As shown in Fig. \ref{Fig:design}B, the projection is conducted via three NN-based encoders as follows.

\begin{figure}[h]
\centerline{\includegraphics[width=0.9\columnwidth]{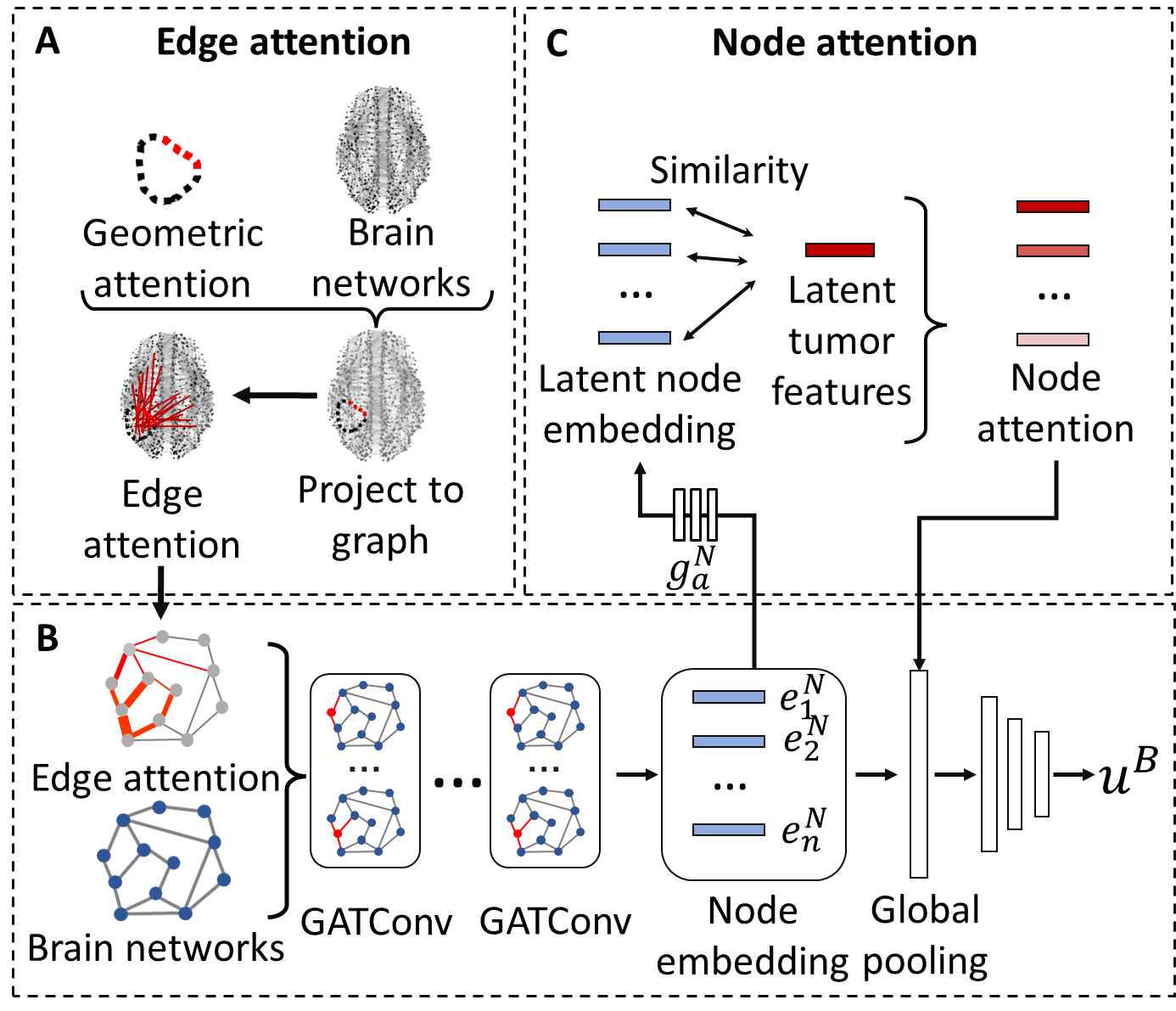}}
\caption{Hierarchical graph attention: A. Geometric boundary attention produced by the geometric encoder is projected to brain networks to obtain edge-level attention. B-C. Edge-attended brain networks are convoluted to produce node embeddings, projected to latent space for generating node-level attention by computing similarity with tumor features. The node-level attention is then utilized in the global pooling level for generating tumor-related brain-level network features $u^B$.}
\label{Fig:architect1}
\end{figure}

\subsubsection{Image encoder} The image encoder is a 3DCNN defined by $u^{I}_i = f^{I} (x^{I}_i)$, where $x^{I}_i$ and $u^{I}_i$ are the image data and output features for the $i$th patient, and $f^{I}(\cdot)$ is the 3DCNN model (see Section~\ref{sec:implement} for implementation details).

\subsubsection{Geometric encoder}
The geometric encoder $f^{P}(\cdot)$ (Fig. \ref{Fig:architect1}A) outputs the geometric features $u^{P}$ and geometric attention $a^{P}$ for every point in the points cloud, defined as $u^{P}_i, a^{P}_i = f^{P} (x^{P}_i)$ for the $i$th patient. 

\subsubsection{Brain network encoder with hierarchical attention}
Brain network features are extracted by training a NN with graph convolution layers, where the NN weights are corrected following a novel hierarchical attention mechanism. 

The attention mechanism is structured by edge-level attention and node-level attention. The former is obtained by projecting the geometric attention of tumor boundary onto the edges (Fig. \ref{Fig:architect1}B). Specifically, the points clouds are projected to the edge atlas. The crossing edges are then assigned with the boundary attention of the points cloud. The edge attention is defined as:
\begin{equation}\label{equ:edgeattention}
a^{E}_{i,j} = \frac{1}{K} \sum_{k}^K ( a^{P}_k) \ ,
\end{equation}
where $a^{E}_{i,j}$ is the edge attention of edge $i,j$. $K$ is the number of points in points cloud crossed by edge $i,j$ and $ a^{P}_k$ is the attention of $k$th point crossed by edge $i,j$.

The outputs of the edge-level attention are further encoded by the GATConv layers that convolute the nodes and edges of the brain networks to obtain a node embedding defined by $e^{N} = f^{B'}(x^{B})$, where $f^{B'}$ is the components of brain network encoder before the global pooling layers (Fig. \ref{Fig:architect1}B). Afterwards, the node embeddings $e^{N}$ are projected to the latent space by a projection head $g^{N}_{a}$. To extract the tumor-related node embeddings, we applied another projection head $g^{T}$ to project the concatenated tumor features $\langle u^{I}, u^{P} \rangle$, composed by both images and points cloud, into the latent space shared with node embedding. We measure the similarity between the node embedding with the tumor features by:
\begin{equation}\label{equ:nodeattention}
    a^{N}_i = S(g^N_{a}(e^{N}), g^{T}(\langle u^{I}, u^{P} \rangle)) \ ,
\end{equation}
where $a^{N}$ is the attention of the $i$th node. $g{a}$ and $g^T_{a}$ are linear projection heads projecting tumor features and node embeddings to the same latent space; $S(\cdot)$ is the similarity function.

By performing both edge and node attention in training the brain network encoder, we extract the most tumor-related features from the brain network and reduce the noise caused by confounding effects, e.g., ageing or other concomitant pathology (Fig.\ref{Fig:architect1}B). The feature extraction of the brain networks is defined as $u^{B}_i = f^{B} (x^{B}_i)$, where $x^{B}_i$ and $u^{B}_i$ are the brain network data and brain network features for the $i$th patient, and $f^{B}$ represents the GNN-based brain network encoder.

\begin{figure}[h]
\centerline{\includegraphics[width=0.8\columnwidth]{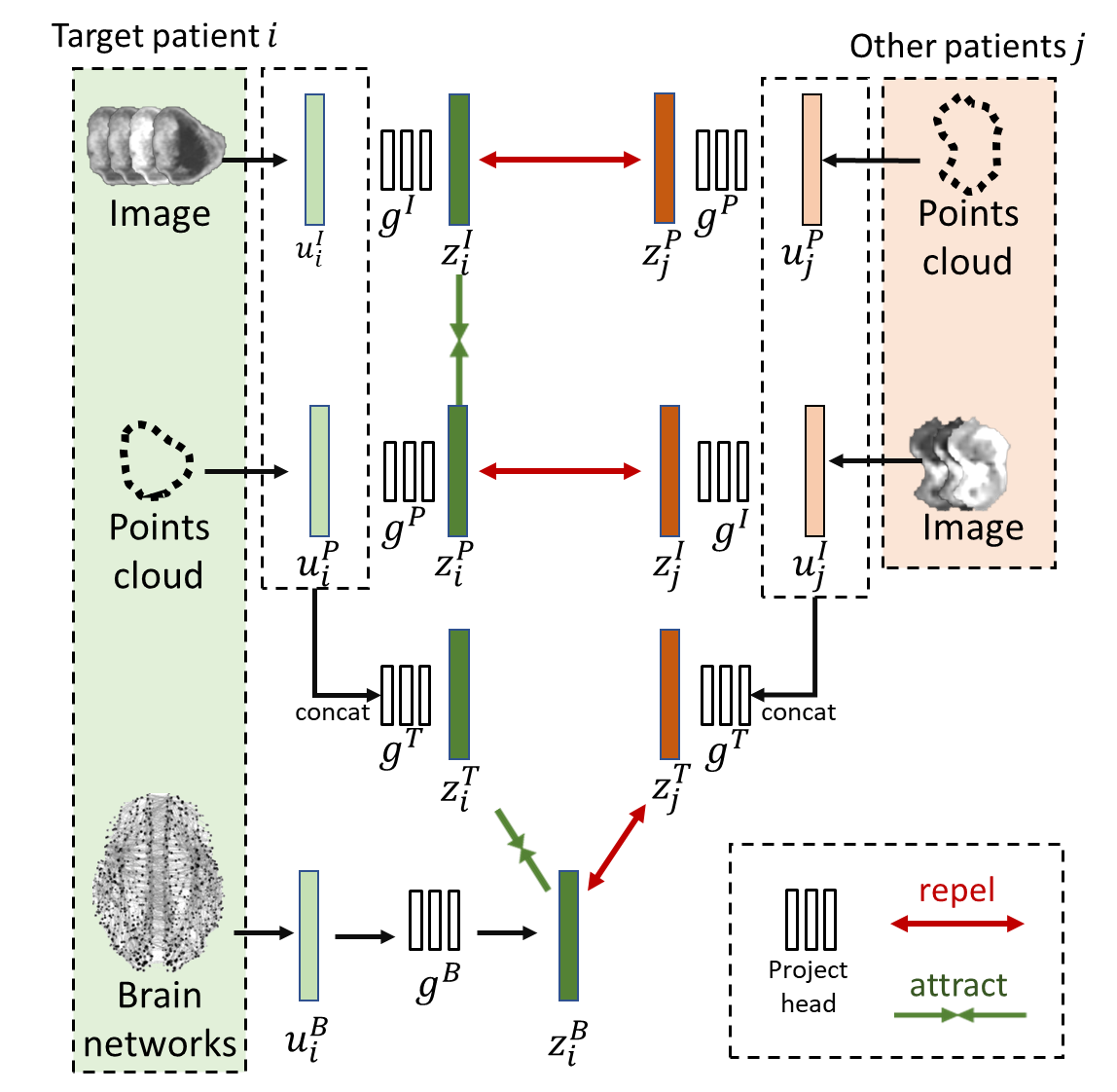}}
\caption{Bi-level multi-modal contrastive learning: Latent features of different modalities $(u^{I}_{i}, u^{P}_{i})$, $(u^{T}_{i}, u^{B}_{i})$ from the same patient (green) attract each other, while latent features of different modalities $(u^{I}_{i}, u^{P}_{j})$, $(u^{T}_{j}, u^{B}_{i})$ from different patients (red) repel each other. The bi-level loss consists of a tumor-level and brain-level components trained together. }
\label{Fig:concept}
\end{figure}

\subsubsection{Bi-level multi-modal contrastive loss}
We develop a bi-level multi-modal contrastive loss to further characterize tumor gradient invasion and minimize the domain gap between the focal tumor and global brain. After extracting the multi-modal features from different encoders, two projection heads are adopted to respectively project the tumor-level features of images and points cloud to the same latent space: $z^{I}_i = g^{I} (u^{I}_i)$ , $z^{P}_i = g^{P} (u^{P}_i)$ where $z^{I}_i$ and $z^{P}_i$ are the projected latent features of images and points cloud, $g^{I}$ and $g^{P}$ are the pre-defined projection heads.

Meanwhile, another two projection heads are employed to respectively project the extracted focal tumor features and brain network features into another latent space: $z^{B}_i = g^{B} (u^{B}_i)$, $z^{T}_i = g^{T} ({u^{I}_i,u^{P}_i})$, where $z^{B}_i$ and $z^{T}_i$ are the projected latent feature of brain networks and focal tumor); $g^{B}$ and $g^{T}$ are the projection head for the brain network and focal tumor.

Subsequently, a bi-level multi-modal contrastive loss is developed to firstly reduce the domain gap of 
tumor-level features by minimizing the cosine distance (attract) between the multi-modal latent features $(z^{I}_{i}, z^{P}_{i})$ from the same patient $i$ and maximizing the cosine distance (repel) of multi-modal latent feature pairs $(z^{I}_{i}, z^{P}_{j})$, $(z^{P}_{i}, z^{I}_{j})$ 
from different patients $i$ and $j$ using the contrastive loss. Secondly, the brain-level domain gap is optimized using a similar approach for the features of brain networks $(z^{B})$ and focal tumor $(z^{T})$.
Due to the asymmetry of the contrastive loss, we design three contrastive losses for tumor image to tumor geometrics (Equation. \ref{equ:I2P}), tumor geometrics to tumor image (Equation. \ref{equ:P2I}) and global brain network to focal tumor (Equation. \ref{equ:B2T}). Finally, we integrate those three sub-losses with a weighting coefficient $\lambda$.
\begin{equation}\label{equ:I2P}
    l^{I2P}_i = -\log \frac{\exp(S(z^{I}_i,z^{P}_i)/\tau)}{\sum_{j\neq i}^{N} \exp(S(z^{I}_i,z^{P}_j)/\tau)} \ ,
\end{equation}
\begin{equation}\label{equ:P2I}
    l^{P2I}_i = -\log \frac{\exp(S(z^{P}_i,z^{I}_i)/\tau)}{\sum_{j\neq i}^{N} \exp(S(z^{P}_i,z^{I}_j)/\tau)} \ ,
\end{equation}
\begin{equation}\label{equ:B2T}
    l^{B2T}_i = -\log \frac{\exp(S(z^{B}_i,z^{T}_i)/\tau)}{\sum_{j\neq i}^{N} \exp(S(z^{B}_i,z^{T}_j)/\tau)} \ ,
\end{equation}
where $i$ is the index of the target patient, and $j$ is the index of other patients in the mini-batch; $S(\cdot)$ is the similar score function; $\tau$ is the temperature parameter; $N$ is the size of the mini-batch. The final multi-modal contrastive loss $\mathcal{L}_{multi}$ is computed as a weighted combination of the above three loss: 
\begin{equation}\label{equ:multicontras}
    \mathcal{L}_{multi}= \frac{1}{N} \sum_{i = 1}^{N} (\lambda (\frac{l^{P2I}_i + l^{I2P}_i}{2}) + (1 - \lambda )l^{B2T}_i ) \ ,
\end{equation}
where $\lambda \in [0,1]$ is a scalar weight coefficient.

\subsubsection{The algorithm}
The proposed multi-modal contrastive learning algorithm is shown in Algorithm \ref{alg1}.

\begin{algorithm}
\caption{Multi-modal contrastive learning}
\label{alg1}
\KwInput{image: $x^{I}$, points cloud: $x^{P}$, brain network: $x^{B}$} 
\For{$i = 1, \cdots$ N}
{
Compute features and attention from image and geometric points cloud: 
$u^{I}_{i} = f^{I} (x^{I})$; $u^{P}_{i}, a^{P} = f^{P} (x^{P})$;  \\
Compute edge attention: $x^{B'}_{i} = x^{B}_{i} \cdot a^{E}$ via ~(\ref{equ:edgeattention}). \\
Compute node embedding using brain network encoder: 
$\{e^{N}_{i,1}, \cdots e^{N}_{i,1}\} = f^{B'}(x^{B}_{i})$. \\

\For{$n = 1, \cdots$ M}
{
Compute node attention $a^{N}_{i,n}$ via ~(\ref{equ:nodeattention}).
}
Extract features from brain networks: $u^{B}_{i} = f^{B} (e^{N} \cdot a^{N})$. \\
Project features to latent space: 
\begin{itemize}
    \item Image: $z^{I}_{i} = g^{I} (u^{I}_{i})$
    \item Points cloud: $z^{P}_{i} = g^{P} (u^{P}_{i})$
    \item Tumor: $z^{T}_{i} = g^{T} (\langle u^{I}_{i}, u^{P}_{i}\rangle)$
    \item Brain networks $z^{B}_{i} = g^{B} (u^{B}_{i}) $
\end{itemize}
Compute multi-modal contrastive loss by ~(\ref{equ:multicontras}).
}
\end{algorithm}
\subsection{Populational graph for classifying glioma patients}

\begin{figure}[t]
\centerline{\includegraphics[width=0.8\columnwidth]{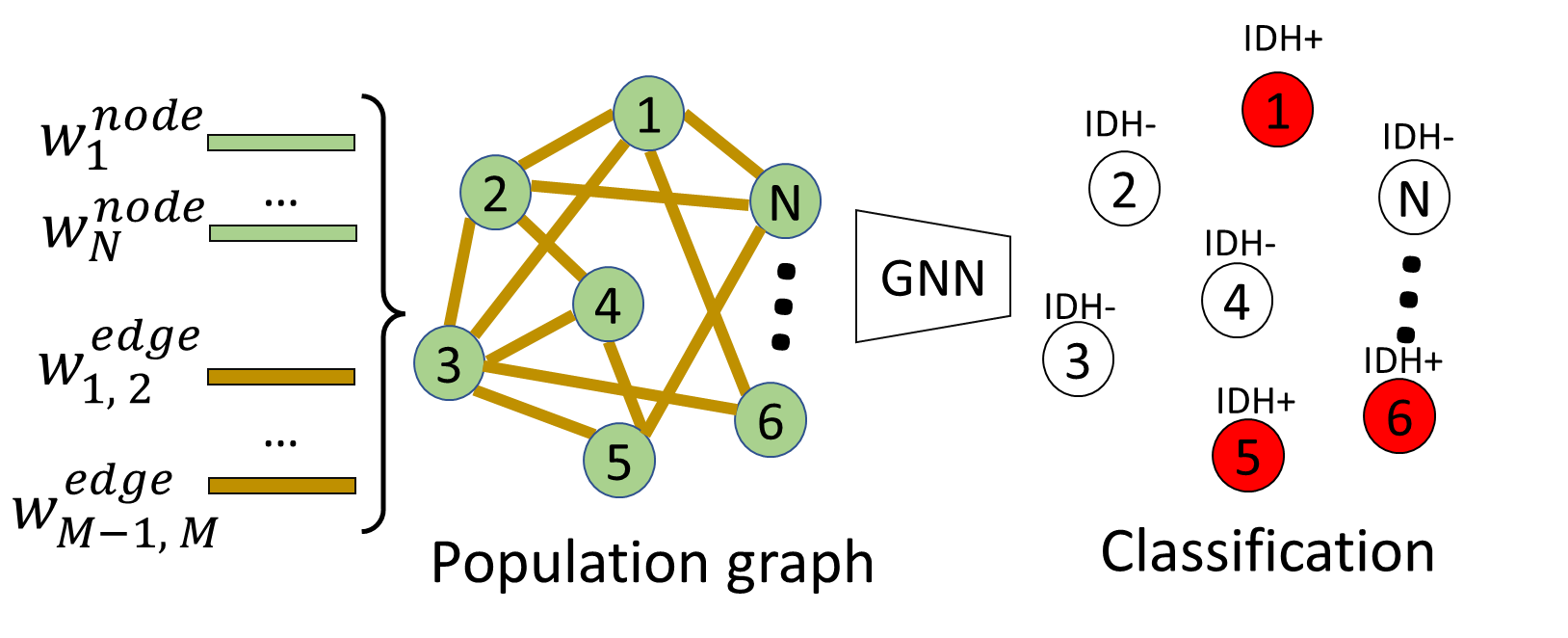}}
\caption{Population graph for patient classification: Each node weight ($w^{node}$) represents features of one patient, while each edge weight ($w^{edge}$) represents the similarity among the features of patients. A GNN node classifier is trained for classifying patients.}
\label{Fig:population}
\end{figure}

With the focal tumor and brain network features generated from the multi-modal learning,
we construct a population graph to characterize the patient cohort (Fig. \ref{Fig:design}C): each node represents the multi-modal features extracted from the patients, while each edge represents the similarity between the multi-modal features among the patients (Fig. \ref{Fig:population}). In the population graph, the node weight of patient $i$ is defined as $w^{node}_{i} = u_i$, and the edge weight between patient $i$ and $j$ is defined as:
\begin{equation}\label{equ:populationedge}
  w^{edge}_{i,j} =
    \begin{cases}
      r(u_i, u_j) \ , & \text{if $r(u_i, u_j) \geq \theta$ }\\
      0 \ , & \text{if $r(u_i, u_j) < \theta$ }
    \end{cases}       
\end{equation}
where $u \in \{u^T, u^B, \langle u^T, u^B \rangle \}$: $u$ is the feature extracted from the multi-modal contrastive learning, and $r(\cdot)$ is the correlation operator. $\theta$ is the threshold of the correlation. 
We design five different combinations of the node weight $w^{node}$ and the edge weight $w^{edge}$ listed in Table \ref{tab:populationgraphcombination}.

\begin{table}[h]
\centering
\caption{Population graph with different node and edge weights}
\label{tab:populationgraphcombination}
\begin{tabular}{cc}
\hline
Node   weight & Edge   weight  \\ \hline
$u^T$ & $r(u^{B})$  \\
$u^B$ & $r(u^{T})$  \\
$\langle u^{T},u^{B} \rangle$ & $r(u^{B})$  \\
$\langle u^{T},u^{B} \rangle$ & $r(u^{T})$  \\
$\langle u^{T},u^{B} \rangle$ & $r(\langle u^{T},u^{B} \rangle)$  \\ \hline
\end{tabular}
\end{table}
Specifically, we consider different combinations of focal tumor features (tumor image and geometrics) and global brain network features as edge and node weight, and we construct the population graph based on the hypothesis that the two categories of features may reflect different patterns of tumor invasion, i.e., localized v.s. widespread invasion. As such, we could integrate two types of features and characterize both the homogeneity and heterogeneity of the cohort.

\section{EXPERIMENTS}
\subsection{Datasets}
We collect the anatomical MRI data of 424 glioma patients available from The Cancer Imaging Archive (TCIA)~\cite{tcgalgg,tcgagbm,ivygap} and an in-house cohort with 117 patients. The MRI modalities include pre-contrast T1, post-contrast T1, T2, and T2-FLAIR. We exclude 17 out of 424 patients due to the low MRI quality or missing IDH mutation status. Finally, 407 of 424 patients are included with 105 IDH mutants and 302 IDH wild-types.

A total of 20 patients of the in-house cohort is used for training the self-supervised models of brain network construction. The remaining 387 patients are split into training and testing set with a 7:3 ratio. The training set of 270 patients is divided by half to train the bi-level multi-modal contrastive learning for feature extraction and population graph-based classifier for patient classification. The testing set includes 117 patients from the publicly available TCIA website. 

\subsection{Image pre-processing}
A standard pre-processing pipeline on MRI data is performed as described~\cite{bakas2017advancing}. Firstly, the pre-contrast T1, T2, and FLAIR images are co-registered to the post-contrast T1 images using the FMRIB's Linear Image Registration Tool of the FMRIB Software Library (FSL)~\cite{jenkinson2012fsl}. Next, skull stripping is performed using the Brain Extraction Tool in FSL~\cite{smith2002fast}. Finally, histogram matching~\cite{nyul2000new} and voxel smoothing with SUSAN noise reduction~\cite{jenkinson2012fsl} are conducted as normalization.

For the in-house cohort with diffusion MRI modalities available, the FA maps are derived from the diffusion MRI using the FMRIB's Diffusion Toolbox. The FA maps are used to train the self-supervised models to extract tract-related features from the anatomical MRI to generate brain networks. 

Finally, all the MRI data are non-linearly transformed to the standard space by co-registering them to the MNI-152-T1-2MM-brain template available in the FSL using the Advanced Normalization Tools~\cite{avants2009advanced}.

TCGA datasets provide manually corrected tumor segmentation masks. 
For other datasets, we utilize the DeepMedic segmentation tool in CaPTk to segment the contrast-enhancing tumor~\cite{kamnitsas2016deepmedic}. Manual correction is performed by a neurosurgeon and a researcher with a DICE score calculated to ensure cross-validation. The image of tumor core is resampled to $2mm\times2mm\times2mm$ and cropped to $120\times120\times120$.

\subsection{Implementation details}
\label{sec:implement}
The proposed framework is tested on an Nvidia 1070 max-Q GPU. All models are implemented using Python. 
\subsubsection{Brain network generation}

The 20 patients yield 46,180 edges and 1,800 nodes for training the self-supervised model. 

For the node autoencoder, all input node voxels are sampled to the dimension of 4,000, the encoder consists of six layers (dimension 2048, 1024, 512, 128, 32, 16), and the output of the bottleneck is a attribute vector $v^{N}$ with a dimension of 16. 
For the edge encoder, all input edge voxels are sampled (Anatomical: 4000, FA: 1000). Two MLP (MRI: 2048, 1024, 512, 128, 32, 16; FA: 1024, 512, 128, 32, 16) respectively encode the voxel vectors of T1 and FA to attribute vectors $v^{E}$ and $v^{E'}$ with a dimension of 16. The final brain networks contain 90 nodes and 2,309 edges with a dimension of 16.
\subsubsection{Image encoder}
The image encoder is a 3DCNN architecture consisting of five 3D convolutional layers (dimension 64, 128, 128, 256, 256), with four input channels corresponding to the four MRI sequences. Batch normalization and max pooling are performed for all convolutional layers. Three feed-forward layers (dimension: 512, 256, 32) are followed to output features with a dimension of 32.

\subsubsection{Geometric encoder}
A specialized GNN is adopted to extract features from the points cloud. The points are first converted into a graph for each convolution by generating links between points and their nearest neighbors within a predefined radius distance. Secondly, convolution operators NNConv~\cite{fey2019fast} aggregate the points features (euclidean coordinates of points) and the link features (distance between points) to the center node. 
Finally, the farthest points sampling is adopted to sample the points with the furthest distance from other points. Our geometric encoder consists of convolutional layers (dimension: 32, 64, 64, 128, 128). After the last layer, a global attention pooling is employed to produce attention scores for the points. Finally, a feed-forward network (dimension: 256, 128, 32) outputs the geometric features with a dimension of 32.

\subsubsection{Brain network encoder}
The projection heads $g^{N}_{a}$ for node attention is NNs (dimension: 16, 32, 64, 128) that projects node embeddings to the latent space shared with focal tumor features $z^{T}$. 
The brain network encoder is a graph attention network with GATConv layers (dimension: 64, 128, 128, 256, 256, 256) that can handle the high-dimensional node and edge attributes~\cite{velivckovic2017graph}. The feed-forward network outputs the brain network features (dimension: 512, 256, 32). Cosine similarity is used as the similarity score $S(\cdot)$ for generating node attention.

\subsubsection{Bi-level multi-modal contrastive loss}
The projection heads $g^I$, $g^P$ and $g^B$, are three separate NNs (dimension 32, 64, 128, 128) that project the features to the latent space with dimension of 128. The projection head for $g^T$ is another NN (dimension 64, 64, 128, 128) that projects the $\langle u^{I}, u^{P} \rangle$ to the latent space with a dimension of 128. 
Cosine similarity is selected as the similarity score $S(\cdot)$. $\tau$ is set to 0.1 $\lambda$ is set to 0.8.

\subsubsection{Population graph and GNN classifier}
$\theta$ of the population graph is set to 0.5. The GATConv is employed as the graph kernel of the GNN to perform node classification in the population graph. The GNN for the population graph consists of layers of GATConv (dimension: 64, 128, 128, 128) followed by pooling layers and a classification layer.

\subsubsection{Training parameters}
For self-supervised learning for generating brain networks, the autoencoder adopts mean squared error loss (MSELoss), the Adam optimizer with a weight decay of 0.0005 and a batch size of 50. We implement the following hyperparameters: 1000 training epochs. We set the initial learning rate as 0.001, and the learning rate is reduced to 90\% after every 50 epochs. For edge reconstruction, we adopt the SGD optimizer~\cite{ruder2016overview} with a weight decay of 0.0005 and a batch size of 50 with 1000 training epochs. We set the initial learning rate as 0.001, and the learning rate is reduced to 90\% after every 50 epochs.

For multi-modal learning, we adopt the SGD optimizer to optimize the network with a weight decay of 0.0005 and a batch size of 20. We implement the following hyperparameters: 1000 training epochs; a mini-batch size of 20. We set the initial learning rate as 0.001, and the learning rate is reduced to 90\% after every 50 epochs. Data augmentation is performed by rotating both images and points cloud data with the same angles.

For population graph-based-GNN, we adopt the Adam optimizer~\cite{ruder2016overview} with a weight a batch size of 20. We apply binary cross-entropy loss for patient classification. We implement the following hyperparameters: 200 training epoch; a mini-batch size of 20. We set the initial learning rate as 0.001, and the learning rate is reduced to 90\% after every 50 epochs.

\subsection{Model evaluation}
\subsubsection{Evaluating performance of the overall framework}
We implement two classic CNN backbones (3D-ResNet34, 3D-DenseNet50) as the benchmark methods of the complete multi-modal learning framework, while the MLP and support vector machines (SVM) are also implemented as the benchmarks to compare the proposed population graph-enhanced GNN.

\subsubsection{Evaluate population graphs}

We conduct experiments in constructing a population graph to choose the best combination of edge and node weights (Table \ref{tab:populationgraphcombination}). 

\subsubsection{Ablation experiments }
We perform ablation experiments to test the importance of different components in the proposed framework. Specifically, we first test the performance of every single encoder of tumor image, tumor points cloud and brain networks, using an MLP and binary cross-entropy loss (BCELoss). Secondly, we test the performance of the pairwise combinations of training two encoders for the classification. Thirdly, we add the contrastive loss $L_{contra}$ to the training of the above two-encoder combinations for multi-task experiments. 
Further, we implement the multi-modal training without the contrastive loss. Finally, we implement the multi-modal training with the hierarchical attention removed.

\subsection{Visualize interpretation}
To interpret the results of the proposed multi-modal learning framework, we identify the critical regions contributing to the prediction from tumor images, tumor points cloud and brain networks using different interpretation approaches.

We employed the Grad-CAM~\cite{selvaraju2017grad} to visualize the critical regions on the tunor images and visualize the geometric attention of the points cloud. To visualize the concordance between points cloud and tumor image, we project the surface points of the Grad-CAM map overlaid on the tumor image to the corresponding points cloud.

To interpret the learning process of the brain network encoder, we employ the GNNExplainer~\cite{ying2019gnnexplainer} to output a probability score that infers the importance of the edges in the brain network. We retain those edges with probability scores greater than $50\%$.

\section{RESULTS}

\subsection{Performance of population graph}

The performance of different approaches of constructing population graph is in Table \ref{tab:populationgraph}. The results show that the population graph achieves the best performance (AUC 0.962) with the concatenated tumor features and brain network features defined as the node and the cosine similarity between brain network features defined as edge. The population graph that uses similarity between tumor features to define edge ($u^B , r(u^T)$: AUC 0.914, $\langle u^T,u^B \rangle , r(u^T)$: AUC 0.940) generally performs worse than those using the similarity between brain network features to define edge ($u^T , r(u^B)$: AUC 0.939, $\langle u^T,u^B \rangle  , r(u^B)$: AUC 0.962). Strikingly, the population graph with concatenated tumor features and brain network features defined as both node and edge performs the worst (AUC 0.888). 

\begin{table}[h]
\centering
\caption{Experiment for selecting the best combination of the population graph}
\label{tab:populationgraph}
\begin{tabular}{cccccc}
\hline
Node   weights & Edge   weights & AUC & ACC & SEN & SPE \\ \hline
$u^T$ & $r(u^{B})$ & 0.939 & 0.879 & 0.894 & 0.812 \\
$u^B$ & $r(u^{T})$ & 0.914 & 0.836 & 0.833 & 0.843 \\
$\langle u^{T},u^{B}\rangle$ & $r(u^{B})$ & \textbf{0.962} & \textbf{0.905} & \textbf{0.917} & \textbf{0.875} \\
$\langle u^{T},u^{B}\rangle$ & $r(u^{T})$ & 0.940 & 0.862 & 0.857 & \textbf{0.875} \\
$\langle u^{T},u^{B}\rangle$ & $r( \langle u^{T},u^{B} \rangle)$ & 0.888 & 0.862 & 0.905 & 0.750 \\ \hline
\end{tabular}
\end{table}

\subsection{Performance of the proposed framework}

\begin{table}[h]
\centering
\caption{Comparing with benmarks}
\label{tab:benchmark}
\begin{tabular}{ccccc}
\hline
Models & AUC & ACC & SEN & SPE \\ \hline
ResNet(3D) & 0.907 & 0.810 & 0.809 & 0.813 \\
DenseNet(3D) & 0.938 & 0.853 & 0.857 & 0.844 \\
Multi-modal + MLP & 0.936 & 0.862 & 0.893 & 0.781 \\
Multi-modal + SVM & 0.932 & 0.871 & 0.893 & 0.813 \\
Multi-modal + GNN & \textbf{0.962} & \textbf{0.905} & \textbf{0.917} & \textbf{0.875}  \\ \hline
\end{tabular}
\end{table}
Our results (Table \ref{tab:benchmark}) show that the best setting of multi-modal framework (AUC 0.962) outperforms the 3D-CNN backbones (DenseNet: AUC 0.938, ResNet: 0.907), implying the importance of the population graph for feature integration. Notably, the performances of  3D-CNN models are higher than the combination of multi-modal contrastive learning with traditional machine learning models (MLP: AUC 0.936, SVM: AUC 0.932).

\subsection{Ablation experiments }
The full results of the ablation experiments are shown in Table \ref{tab:Ablation}. The experiments of the individual encoder show that the brain encoder (AUC 0.877) outperforms the tumor geometric (AUC 0.858) and tumor image (AUC 0.869) encoders. The multi-task two-encoder experiments indicate that the best combination of data modalities is tumor image and tumor points cloud (AUC 0.874). The two-encoder experiments with the additional contrastive loss show that combining tumor image and tumor geometric encoders (AUC 0.929) consistently performs the best. Finally, removing the graph attention modules significantly decreases the performance of the contrastive training framework (with attention: AUC 0.936, without attention: AUC 0.924).

\begin{table}[]
\centering
\caption{Ablation experiments }
\label{tab:Ablation}
\begin{tabular}{ccccc}
\hline
Ablation experiments & AUC & ACC & SEN & SPE \\ \hline
$f^{I}$+ MLP & 0.877 & 0.802 & 0.810 & 0.781 \\
$f^{P}$+ MLP & 0.858 & 0.813 & 0.809 & 0.812 \\
$f^{B}$+ MLP & 0.869 & 0.810 & 0.821 & 0.781 \\
$f^{I}+f^{P}$+ MLP & 0.874 & 0.828 & 0.833 & 0.812 \\
$f^{B}+f^{P}$+ MLP & 0.871 & 0.819 & 0.833 & 0.781 \\
$f^{I}+f^{B}$+ MLP & 0.876 & 0.836 & 0.860 & 0.750 \\
$f^{I}+f^{P}$+ MLP+ $L_{contr}$ & 0.929 & 0.853 & 0.893 & 0.750 \\
$f^{B}+f^{P}$+ MLP+ $L_{contr}$ & 0.887 & 0.836 & 0.869 & 0.688 \\
$f^{I}+f^{B}$+ MLP+ $L_{contr}$ & 0.894 & 0.845 & 0.833 & \textbf{0.875} \\
Multi-modal (no attention) + MLP   & 0.924 & 0.853 & 0.869 & 0.813 \\
Multi-modal (no $L_{multi}$) + MLP & 0.910 & 0.836 & 0.833 & 0.844 \\
Multi-modal (full model)  + MLP & \textbf{0.936} & \textbf{0.862} & \textbf{0.893} & 0.781 \\ \hline
\end{tabular}
\end{table}

\begin{figure}[h]
\centerline{\includegraphics[width=0.8\columnwidth]{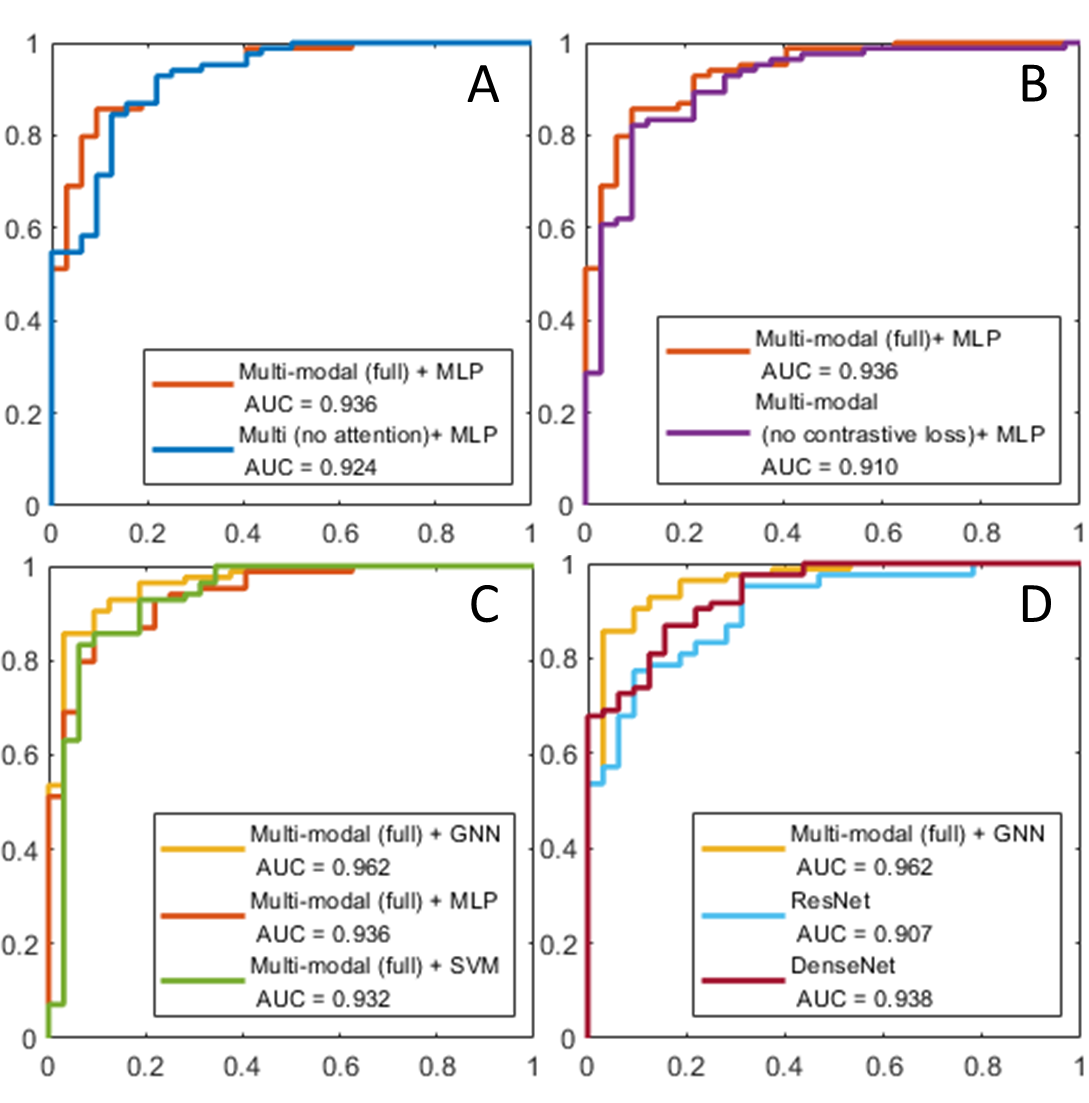}}
\caption{AUC for the ablation experiments: A. Models with and without hierarchical attention. B. Models with and without bi-level contrastive loss. C. Models of the full framework and the benchmarks of population graph-enhanced GNN. D. Models of the full framework and CNN benchmarks.}
\label{Fig:concept}
\end{figure}

\subsection{Interpretative visulization}
\begin{figure}[h]
\centerline{\includegraphics[width=0.8\columnwidth]{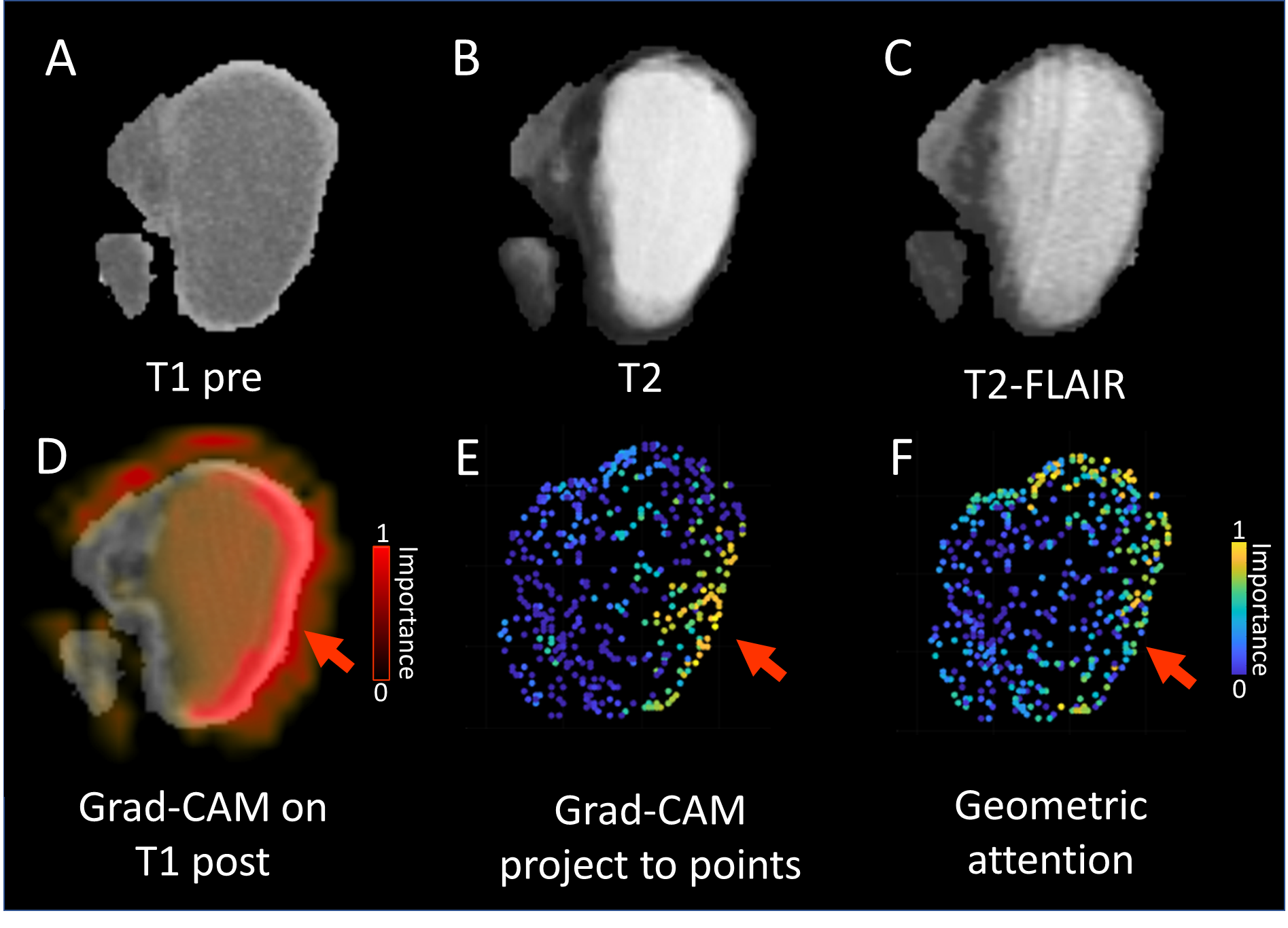}}
\caption{An case example of interpretation of image and geometric encoders. A. pre-contrast T1; B. T2; C. T2-FLAIR. D. The Grad-CAM heatmap overlaid on post-contrast T1. E. The Grad-CAM voxels projected to the points cloud. E. Points attention generated by the geometric encoder.}
\label{Fig:shapeimage}
\end{figure}

\begin{figure}[h]
\centerline{\includegraphics[width=0.8\columnwidth]{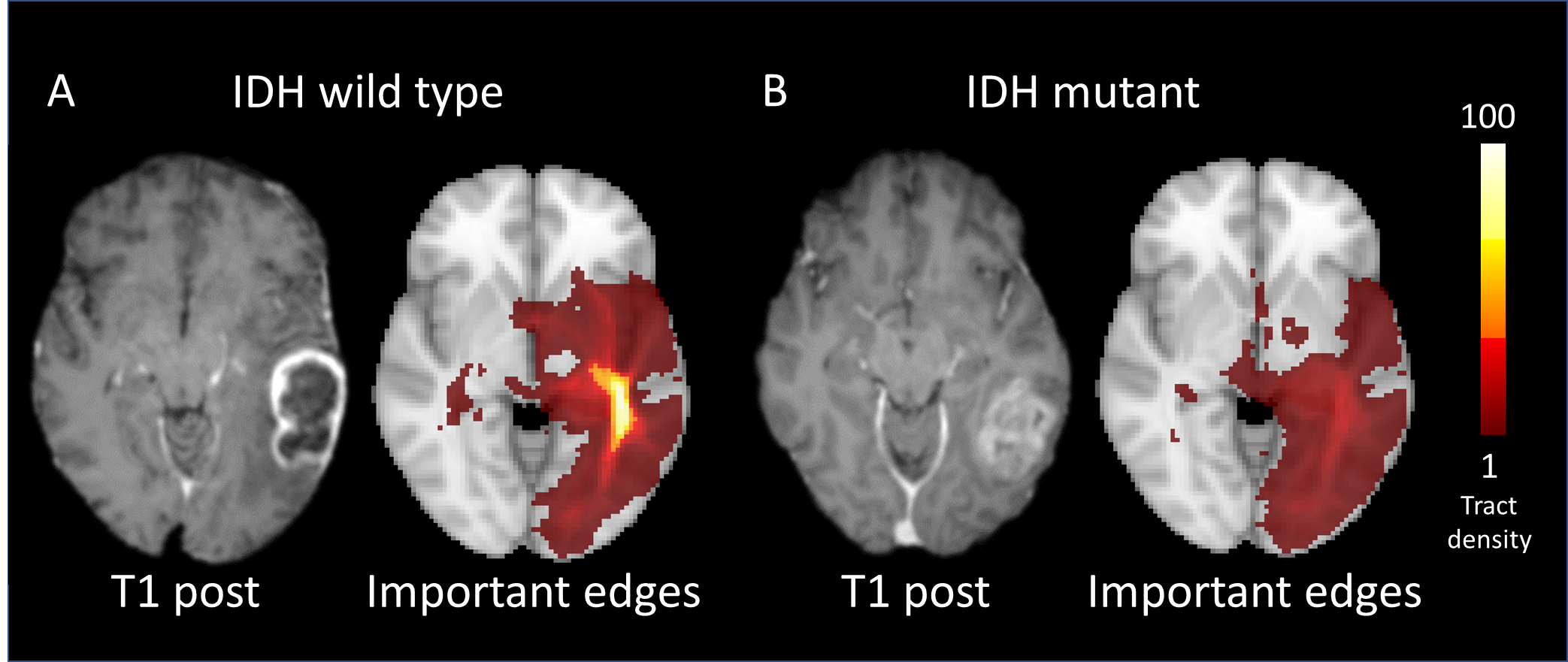}}
\caption{Examples of IDH mutant and wild-type. A IDH wild-type. B. IDH mutant. Voxel distribution of disrupted tracts with over $50\%$ probability of importance are indicated. }
\label{Fig:brainnetvis}
\end{figure}

The interpretative visualizations in Fig.\ref{Fig:shapeimage} show that the image encoder and the geometric encoder indicate common tumor regions (Fig.\ref{Fig:shapeimage}D - F) important for model prediction, which suggests that these regions are specific to IDH mutation. Fig.\ref{Fig:shapeimage}D shows that the Grad-CAM focuses on the tumor contrast-enhancing edges with high intensity in both T2 and T2-FLAIR images (Fig.\ref{Fig:shapeimage}B, C). Through the visualization in Fig.\ref{Fig:brainnetvis}, we find that the brain networks of IDH wild-type demonstrate a higher density of important disrupted edges compared to IDH mutant. This finding aligns with our prior knowledge that the IDH wild-type is generally more invasive than the IDH mutant.

\section{DISCUSSION}
We propose a multi-modal contrastive learning framework that exploits the multi-modal features extracted from the tumor image, points cloud and global brain networks for predicting glioma genotype. We firstly develop a novel self-supervised learning approach to construct brain networks from anatomical multi-sequence MRI. Moreover, tumour-related brain network features can be extracted by developing hierarchical graph attention for the brain network encoders. Further, we design a bi-level multi-modal contrastive loss that could align tumor-related network features with focal tumor features across the domain gap. Finally, we construct a population graph that could effectively integrate the multi-modal features and predict patients' genotype. Our learning framework achieves the highest performance compared to benchmark methods.



Previous studies show that the tumor geometric features demonstrate crucial value in characterizing tumors. Although the single geometric encoder does not perform the best in our experiments, the combination of image and geometric encoders shows high performance, which suggests the benefit of including geometric data to enhance extractingthe most relevant features. The interpretative visualization shown in Fig.\ref{Fig:shapeimage} indicates the agreement between points cloud and image features, which could further validate the effectiveness of the multi-modal contrastive learning in aligning multiple modality domains. Biologically, this could be interpreted as the association between tumor content and tumor boundaries, indicating tumor aggressiveness and 
invading patterns. 

Glioma is characterized by diffuse infiltration beyond the focal lesion, disrupting the global brain. Therefore, characterizing the brain network promises to add value to the focal tumor.
However, glioma patients also frequently demonstrate concomitant pathology beyond the lesion, which could challenge extracting specific features from the brain network for predicting the genotype. 
To efficiently extract tumor-specific features from the brain network, we introduce a hierarchical graph attention that attends to the edges of the brain network associated with tumor points cloud, indicating the white matter tracts (Fig. \ref{Fig:brainnetvis}) responsible for tumor invasion across tumor boundaries (Fig. \ref{Fig:shapeimage}). 
Similarly, the attention module could further identify the nodes associated with the focal tumor. This neuroscience-inspired attention module demonstrates significance in enhancing the model performance and interpretability, shown by the ablation experiments and visualization.

Our experiments show that the multi-modal contrastive learning approach outperforms the CNN-based benchmarks, validating the usefulness of properly incorporating tumor geometrics and brain networks in predicting glioma genotype. Our hierarchical attention transfer the geometric attention of points cloud to the crossing edges and minimize the domain gap between focal tumor and brain networks. Different from the traditional cross-modal attention~\cite{lee2018stacked}, our hierarchical attention is designed based on the data properties and clinical hypothesis. In addition, we develop a bi-level contrastive loss, tailored to perform contrastive learning at tumor and brain levels, reflecting the gradient invasion pattern. Instead of directly applying cross-modal contrastive loss between three modalities in the same latent space, we design a bi-level approach that considers the tumor gradient effect across the brain.

To integrate the multi-modal features, we develop a population graph to characterize the patient cohort. The brain network features demonstrate as the best features describing patient similarity, 
the importance of network features. In contrast, focal tumor features show weaker performance in characterizing patient similarity, which might be due to the remarkable tumor heterogeneity and the limited information compared to the global brain, further supporting the value of incorporating comprehensive features in the prediction. 



The proposed methods have the potential for automated, rapid diagnosis and prognosis in glioma patients based on pre-treatment MRI, which is essential for patient risk stratification and treatment planning towards precision medicine. Further, our hierarchical graph attention could help reveal the tumor-related disruption beyond the lesion, which could help enhance more precise planning of surgery and radiotherapy, as recent studies show that identifying disrupted white matter tracts could help reveal invisible tumor invasion on the conventional MRI and indicate the recurrence location~\cite{wei2021structural}. 

This study has limitations. Firstly, due to the rarity of glioma, our training sample is smaller than other cancers, although our cohort is one of the largest in glioma. Secondly, our cohort is slightly more imbalanced ($\sim 25\%$ IDH mutant) than reported incidence ($\sim 40\%$)~\cite{liu2019gene}. We use both AUC and accuracy in evaluating our model performance, which shows comparable results, implying the model robustness. Thirdly, constructing brain networks relies on the neuroanatomy atlases, where we use an atlas with 90 brain regions, due to the limitation of computational costs. Adopting the atlases with higher resolution could further increase the framework's performance. Our future work will involve larger datasets and transfer learning to further enhance the performance. In addition, manifold or mesh-based geometric encoder could be utilized to capture features from a more detailed geometric data format.

\section{CONCLUSION}
We present a novel multi-modal learning framework for predicting the IDH mutation of glioma. Our technical contribution include: a self-supervised approach for generating brain networks from anatomical MRI; a specialized hierarchical attention module that attends to tumor related edges and nodes; a bi-level contrastive loss for minimizing the domain gap between different modalities; a weighted population graph for feature integration and patient classification. Our framework outperforms the classic CNN backbones, while the population graph-based classifier outperforms traditional machine learning models. In future, we will further develop our model to include clinical variables into the prediction model.

\bibliographystyle{splncs04}
\bibliography{refs.bib}

\end{document}